\pdfoutput=1

\documentclass[11pt]{article}

\usepackage[]{acl}

\usepackage{times}
\usepackage{latexsym}

\usepackage[T1]{fontenc}

\usepackage[utf8]{inputenc}

\usepackage{microtype}

\usepackage{url}
\usepackage{times}
\usepackage{paralist,tabularx}
\usepackage{latexsym}
\usepackage{multirow}
\usepackage{graphicx}  
\usepackage{amsmath}
\usepackage{multicol}
\usepackage{amssymb}
\usepackage{booktabs}
\usepackage{color}
\usepackage{colortbl,array,xcolor}
\usepackage{xspace}
\usepackage{wrapfig}

\newcommand{\ie}{i.e.,~}
\newcommand{\eg}{e.g.,~}

\newcommand{\Ni}{({\em i})~}
\newcommand{\Nii}{({\em ii})~}

\definecolor{cyan}{rgb}{0.88,1,1}

\newcommand{\simo}[1]{\textcolor{black}{#1}}

\newcommand{\gc}[1]{\textcolor{black}{#1}}

\usepackage{cleveref}
\usepackage{collcell}
\usepackage{color, colortbl}
\usepackage{algorithm}
\usepackage{algpseudocode}
\usepackage{subcaption}

\title{Evaluation Metrics of Language Generation Models \\ for Synthetic Traffic Generation Tasks}

\author{Simone Filice\textsuperscript{2}, Jason Ingyu Choi\textsuperscript{1}, Giuseppe Castellucci\textsuperscript{1}, Eugene Agichtein\textsuperscript{1}, Oleg Rokhlenko\textsuperscript{1} \\
     \textsuperscript{1}Amazon, Seattle - USA \\
     \textsuperscript{2}Amazon, Tel Aviv - Israel\\
     \texttt{\{filicesf,chojson,giusecas,eugeneag,olegro\}@amazon.com}}

\begin{document}
\maketitle

\begin{abstract}
Many Natural Language Generation (NLG) tasks aim to generate a single output text given an input prompt. Other settings require the generation of multiple texts, \eg for Synthetic Traffic Generation (STG). This generation task is crucial for training and evaluating QA systems as well as conversational agents, where the goal is to generate multiple questions or utterances resembling the linguistic variability of real users.
In this paper, we show that common NLG metrics, like BLEU, are not suitable for evaluating STG. We propose and evaluate several metrics designed to compare the generated traffic to the distribution of real user texts. 
We validate our metrics with an automatic procedure to verify whether they capture different types of quality issues of generated data; we also run human annotations to verify the correlation with human judgements. Experiments on three tasks, \ie Shopping Utterance Generation, Product Question Generation and Query Auto Completion, demonstrate that our metrics are effective for evaluating STG tasks, and improve the agreement with human judgement up to 20\% with respect to common NLG metrics. \simo{We believe these findings can pave the way towards better solutions for estimating the representativeness of synthetic text data.}
\end{abstract}

\section{Introduction}
\label{sec:intro}

Synthetic Data Generation (SDG) is often used to augment the training material of Natural Language Processing (NLP) models \cite{DBLP:journals/corr/abs-2105-03075}. Synthetic data is needed as the increasing complexity of NLP models makes them data hungry, while privacy concerns complicate the acquisition, storage and annotation of real data.
SDG is particularly useful for AI assistants, since large-scale data is needed to train and to track their performance. Text generation is controlled by prompting the
model with the content to verbalize. 
For example, to generate shopping utterances for voice-based e-commerce, the input can include an intent, \eg search, and slotting information, \eg a product. Given the multitude of linguistic expressions for searching a product, 
NLG models must generate multiple outputs for the same prompt. We refer to this single-prompt-multi-output setting as Synthetic Traffic Generation (STG).

\gc{Evaluating NLG models for STG is an open question}. Common solutions, \eg BLEU \cite{papineni-etal-2002-bleu}, ROUGE \cite{lin-2004-rouge}, or BERT-score \cite{DBLP:conf/iclr/ZhangKWWA20}, independently \gc{rate} each text. As shown in table \ref{tab:examples}, averaging per-utterance scores is not ideal. The table compares synthetic and user utterances having the same \textit{search} intent about running shoes by Nike; each synthetic utterance is individually good, but 
if we consider entire bags\footnote{A \textit{bag} refers to a set of utterances with repetitions, which allows to distinguish between frequent and rare expressions.}, it is clear that the generated data does not resemble real traffic.

\begin{table*}[!ht]
\centering
\begin{tabular}{c|c}
\textbf{Real Traffic Data}                  & \textbf{Synthetic Traffic Data}                \\
\hline
Search for nike running shoes       & Search for nike running shoes \\
Look for shoes for running          & Search for nike running shoes \\
Do you have running shoes from nike & Search for nike running shoes \\
Search nike shoes                   & Search for nike running shoes \\
Can you show me blue running shoes  & Search for nike running shoes \\
\end{tabular}
\caption{A model generating individually good utterances is not necessarily good in single-prompt-multi-output settings. All utterances in the table have been manually created. For privacy concerns we do not report any real user data.}
\label{tab:examples}
\end{table*}

In this paper, we propose several metrics to evaluate NLG models for STG. Our metrics perform a bag-level comparison between generated texts and real user data. To validate our metrics, we design an automatic procedure where the reference bag is manipulated using different types of noise. We compare the resulting noisy bags with the original bag and verify whether our metrics can capture synthetically introduced noises. We further conduct manual assessments to verify the correlation between the metrics and human judgments on deciding which generated bag is more similar to the reference one. Experiments using one publicly available dataset and two real industry scenarios show that our proposed bag-level metrics are superior to standard NLG metrics that average all possible pairwise scores. \simo{Nevertheless, evaluating the quality of synthetic data is still an open problem that deserves special attention from the community.} 
From our knowledge, this is the first work that studies a wide range of existing metrics in the context of STG, \simo{and we believe our findings represent a valuable starting point in this research direction}.

In the rest of the paper, \cref{sec:related} reports the related works. \cref{sec:metrics} and \cref{sec:experiments} describe the proposed metrics and the experiments, respectively. Finally, \cref{sec:conclusions} discusses the conclusions.

\section{Related Work}
\label{sec:related}



Evaluation in NLG is challenging as many tasks are open ended and there are almost infinite ways to express a concept \cite{DBLP:journals/corr/GattK17}. Human judgement is the gold standard but it is expensive and time-consuming;
researchers thus often resort to automatic metrics.
Common metrics are untrained and evaluate the n-gram overlap between generated and reference texts. For example, Bilingual Evaluation Understudy (BLEU) \cite{papineni-etal-2002-bleu}, often used in Machine Translation, computes the weighted geometric mean of n-gram precision scores; Recall-Oriented Understudy for Gisting Evaluation (ROUGE) \cite{lin-2004-rouge}, initially proposed in automatic summarization, focuses on recall; Consensus-based Image Description Evaluation (CIDEr) \cite{DBLP:journals/corr/VedantamZP14a}, proposed for image captioning, uses tf-idf to compute the  weighted n-gram overlap. Others relax the lexical match by using synonyms (\eg Metric for Evaluation of Translation with Explicit ORdering (METEOR) \cite{banerjee-lavie-2005-meteor}) or embeddings similarity (\eg MoverScore \cite{DBLP:journals/corr/abs-1909-02622}).

Other metrics are machine learned: BERTscore \cite{DBLP:journals/corr/abs-1904-09675} uses BERT embeddings \cite{devlin-etal-2019-bert} to match candidate words by cosine similarity; Sentence-BERT (SBert) \cite{DBLP:journals/corr/abs-1908-10084} is a Siamese network to compute the cosine similarity between BERT sentence embeddings; BLEURT \cite{sellam-etal-2020-bleurt} is a BERT model fine-tuned to provide human-like ratings. 

The above metrics compare a generated text with a reference. Since a single reference cannot cover all the plausible outputs, researchers propose to use multiple references to improve the correlation with human judgments \cite{3178deffe35547908a650aff00260140}. Some metrics, \eg BLEU, support multi-references, while others can be extended by computing the average or max score across all references. This single-generation-multi-reference comparison is still different from our use case, as we need to compare multiple generated outputs to multiple references. 


In the context of Generative Adversarial Networks \cite{Goodfellow2014GenerativeAN}, some metrics have been proposed to compare distributions of generated and reference images \cite{BORJI201941}. These are tailored for the Computer Vision domain and cannot be easily applied to NLG. For a more detailed survey on NLG evaluation, please refer to \citet{DBLP:journals/corr/abs-2006-14799}. 


\section{Metrics for Synthetic Traffic Generation}
\label{sec:metrics}

\begin{table*}[!ht]
\centering
\begin{tabular}{c|c}
\textbf{Metric} & \textbf{Description}\\
\hline
Pair$_{\text{BLEU-3}}$ & Averaged pairwise scores using BLEU-3. \\
Pair$_{\text{ROUGE-L}}$ & Averaged pairwise scores using ROUGE-L. \\
Pair$_{\text{CIDEr}}$ & Averaged pairwise scores using CIDEr. \\
Pair$_{\text{SBert}}$ & Averaged pairwise scores using SBert. \\
Cos$_{\text{TF}}$ & Cosine similarity of TF representations of $G$ and $R$.\\
Cos$_{\text{TF-IDF}}$ & Cosine similarity of TF-IDF representations of $G$ and $R$.\\
Clus$_{\text{TF}}$ & DBSCAN applied to TF encodings of $G$ and $R$.\\
InvPP & Inverse of perplexity of 4-gram language model. \\
InvKL & Inverse of KL divergence of unigram distributions. \\
Align$_{\text{BLEU-3}}$ & Alignment-based metric using BLEU-3.\\
Align$_{\text{ROUGE-L}}$ & Alignment-based metric using ROUGE-L.\\
Align$_{\text{CIDEr}}$ & Alignment-based metric using CIDEr.\\
Align$_{\text{SBert}}$ & Alignment-based metric using SBert.\\
\end{tabular}
\caption{Metrics used in the experimental evaluations.}
\label{tab:metrics_evaluation}
\end{table*}

We propose different families of metrics. 
In the following, we refer to the generated and reference bags with $G$ and $R$, respectively.\\

\noindent
\textbf{Pairwise Metrics}.
A na\"ive solution for estimating the bag-to-bag similarity is computing the average sentence-to-sentence similarity between all the pairs from the two bags. 
More formally, given a sentence-to-sentence similarity metric $sim$, we define the pairwise bag-to-bag similarity as:

\begin{equation*}
    Pair_{sim}(G, R)=\frac{\sum_{g \in G, r \in R}sim(g, r)}{|G| |R|}
\end{equation*}


\noindent This solution tends to favor generated bags that contain mostly texts from the head of the reference distribution (\ie the most frequent expressions). The reason is that each text in $G$ is compared to each text in $R$, and head texts maximize the average similarity.\\




\begin{algorithm*}[ht]
\footnotesize
\caption{Bag Similarity by Clustering}\label{alg:cap}
\begin{algorithmic}
    \Require $G, R, \mathcal{E}$
    \Ensure Similarity score
    \State $B \gets G \cup R$ \Comment{Combine two bags and keep duplicates}
    \State $p(B) \gets \frac{|R|}{|B|}$ \Comment{Expected percentage of texts from $R$}
    \For{text $u$ in $B$}
        \State compute $\mathcal{E}(u)$ \Comment{Encode each sentence}
    \EndFor

    \State $\mathcal{K}\gets$ run DBSCAN to cluster vectors $\mathcal{E}(u)$  \Comment{Fit clustering}
    \For{cluster $C$ in $\mathcal{K}$} 
        \State $p(C)$ $\gets \frac{|C \cap R|}{|C|}$ \Comment{percentage of texts from $R$ in cluster $C$}
        \State $d(C)$ $\gets |p(B) - p(C)|$ \Comment{difference from expected percentage of texts from $R$}
        \State $\bar{d}(C) \gets d(C) \cdot \frac{|C|}{|B|}$) \Comment{Weight the difference by cluster size}
    \EndFor\\
    \Return $cluster_{\mathcal{E}} = \frac{1}{\sum_{C}\bar{d}(C)}$ \Comment{Return inverse of the weighted average of the differences}
\label{clustering_pseudocode}
\end{algorithmic}
\end{algorithm*}

\noindent
\textbf{Alignment-based Metrics}.
Word alignment has been extensively studied in machine translation \citep{alignment1, alignment2}. 
We propose metrics based on sentence-level alignment. In particular, we expand the ideas proposed in graph algorithms \citep{bhagwani2012sranjans} by representing $G$ and $R$ as a bipartite graph where each sentence from $G$ and $R$ corresponds to a node. We create edges $(g,r)$ connecting each node $g \in G$ to each node $r \in R$ by assigning a weight as $sim(g,r)$, where $sim$ can be any existing sentence-to-sentence similarity metric. 
To compute sentence-level alignments, we apply an existing maximal matching algorithm \citep{blossom} to the resulting graph and obtain the sentence-level alignment $A(G,R)$. Once maximal matching pairs are found, we compute the bag-to-bag similarity as:

\begin{equation*}
    Align_{sim}(G, R)= \frac{\sum_{(g, r) \in A(G, R)}sim(g, r)}{|G|}
\end{equation*}

\noindent This is essentially summing the weights that maximize the pairwise similarity defined by any sentence similarity metric, normalized by the length of two bags\footnote{In this formulation we are assuming that $|R|=|G|$. If instead the number of texts in the two bags are different, they can be made equal by using upsampling or downsampling.}.
In our formulation we enforce a strict 1-to-1 alignment, \ie each node from $G$ is aligned to a single node from $R$, and vice versa. Note that if there are duplicate texts in a bag, we simply create multiple copies of the same node. \\


\noindent
\textbf{Clustering Metrics}.
We explore metrics proposed for data clustering, such as cluster purity, which measure how balanced class labels are within each cluster. Specifically, given $R$ and $G$ and any sentence encoder $\mathcal{E}$, we  estimate the bag-to-bag similarity using the procedure illustrated in Algorithm \ref{alg:cap}.
We mix $R$ and $G$ into a bag $B$ and measure $p(B)$ as the percentage of texts from $R$ in $B$. Then, we apply DBSCAN \cite{ester1996density} to $B$. If $R$ and $G$ are similarly distributed, the resulting clusters should contain texts of both bags, otherwise the clusters should have a higher purity, \ie containing texts only from $R$ or $G$. For each cluster $C$ we can compute the difference between its percentage of texts from $R$, namely $p(C)$, and the expected percentage $p(B)$. The bag similarity is the weighted average of these values.
We use DBSCAN as it does not need to specify the number of clusters: indeed, the optimal number of clusters is unknown. Intuitively, this value corresponds to the number of sub-modalities users can adopt to verbalize a given concept.\\



\noindent
\textbf{Document Similarity Metrics}.
We consider also document similarity solutions: given a sentence encoder $\mathcal{E}$, we compute the vector representation $\vec{B}$ of a bag $B$ by summing up the encoding of its texts, \ie $\vec{B} = \sum_{u \in B}\mathcal{E}(u)$. The similarity between $R$ and $G$ is then the cosine similarity of their vectors:

\begin{equation*}
    Cos_{\mathcal{E}}(G, R) = \frac{\vec{G} \cdot \vec{R}}{\| \vec{G} \|\| \vec{R} \|}
\end{equation*}

\noindent We also consider representing the bags as their uni-gram probability distribution and compute the Kullback–Leibler divergence \citep{joyce2011kullback} $D_{KL}(G||R)$. As a similarity score, we adopt the inverse of such value:

\begin{equation*}
    InvKL(G, R) = D_{KL}(G||R)^{-1}
\end{equation*}

\noindent
\textbf{Language Model Metrics}.
We define a metric inspired by the ASR and language model literature. We train a language model\footnote{We adopt a 4-gram language model with Knerser-Ney Smoothing \cite{DBLP:conf/interspeech/HsuG08}.} using $G$ and compute the perplexity of texts in $R$, \ie $PP_G(R)$. The final score is then the inverse of the perplexity: 

\begin{equation*}
    InvPP(G, R) = PP_G(R)^{-1}
\end{equation*}


\section{Evaluating the Evaluation Metrics}
\label{sec:experiments}

In this section we describe two strategies - one entirely automatic, the other one based on human judgements - to validate and identify the most promising metrics for STG. 
Refer to Table \ref{tab:metrics_evaluation} for a summary of the metrics we adopt in the experiments below.

\subsection{Evaluation Tasks and Data}
\label{sec:eval_tasks}


\noindent
\textbf{Product Question Generation (PQG)}. Given a product we aim to generate product related questions. We prompt a NLG model with the product title, product category and product attribute type (\eg shoes type, hard drive capacity).
We adopt two open-source datasets: Amazon Product Question Answers (Amazon-PQA) \citep{amazon_pqa} and MAVE \citep{yang2022mave}. The former contains 10M product questions/answers from \texttt{amazon.com}. The latter contains product category and product attribute type-value annotations on 2.2M Amazon products. We select the product questions from Amazon-PQA corresponding to products in MAVE. We apply keyword matching to identify the questions containing category-specific attribute values. For example, the question ``\textit{How many mb is in the 64 gb?}'' contains the value ``\textit{64 gb}'' for the attribute \textit{``usb flash drives capacity''}. By applying this procedure we obtained 84,044 questions that contain product category/attribute annotations from MAVE. Following the context $C$ definition from Section \ref{sec:eval_tasks}, there are 31,727 unique contexts distributed across 22,900 unique products. There are 1,246 (3.9\%) contexts that contain 10+ questions, 9,982 (31.4\%) with 2-9 questions and 20,499 (64,6\%) only have 1 question. \\

To create a test split, we sampled 1,000 contexts from 10+ questions group since we had to ensure test samples contain at least 10 questions. Similarly for development set, 1,000 contexts are sampled from 2-9 questions group. Lastly, all remaining 29,727 contexts are allocated to training set. There are 55.8k, 3.2k and 24.8k questions in training, development and test sets, respectively.

\noindent
\textbf{Shopping Utterance Generation (SUG)}. Given a product and an intent we aim to generate voice shopping utterances for a conversational assistant. We consider \textit{buy, search, add to cart} and \textit{check price} intents.
To create the data for SUG, we used 13 months of logs from the real traffic of a shopping assistant, from which we extracted de-identified (anonymous) utterances, along with their intent and the purchased/searched product. The data from the first 12 months have been used for training and the remaining for evaluation.

\noindent
\textbf{Query Auto Completion (QAC)}. Given a product search query we aim to generate query auto-completions. We collected 50k train and 5k test queries from our search logs. The reference bags include the top-10 queries obtained from the Amazon auto-completion API\footnote{\texttt{https://completion.amazon.com}}.

\subsection{Automatic Evaluation}
\label{sec:automatic_evaluation}

We propose a scalable automatic evaluation procedure. Starting from a reference bag $R$, we create a ranking of multiple generated bags $\mathcal{R}^* = [G_1, G_2, ..., G_n]$ byincrementally applying multiple manipulations (or by applying the same manipulation with increased strength). This procedure guarantees that the bags in $\mathcal{R}^*$ are always ranked by their level of noise.  We use a metric $m$ to compute the similarity between $R$ and each $G_i$. Finally, we rank the generated bags according to the metric scores, and verify whether the resulting ranking $\mathcal{R}_m$ correlates with the real ranking $\mathcal{R}^*$. Our manipulations include:

\begin{itemize}
    \item \gc{\textbf{Text Distribution Manipulation (TDM)}} (not in QAC): we alter the original distribution in $R$ to be more peaked (\ie we substitute occurrences of tail texts with head ones) or flatter (\ie we equalize the occurrences of each text in the bag).
    \item \gc{\textbf{Noisy Text Injection (NTI)}}: we replace an increasing number of texts of $R$ with texts having different intents (SUG), products (SUG and PQG), or completions of different queries (QAC). 
    \item \gc{\textbf{Easy Data Augmentation (EDA)}} \cite{DBLP:journals/corr/abs-1901-11196}: we modify an increasing number of texts from $R$ by applying word swapping, replacement, deletion or insertion.
    \item \gc{\textbf{Carrier Phrase}}\footnote{With \textit{carrier phrase} we indicate the sentence portion indicating the user intent, \eg \textit{I want to buy} in the sentence \textit{I want to buy the latest iphone}.} \gc{\textbf{Substitution (CPS)}} (only for SUG): we modify an increasing number of texts in $R$ by changing their carrier phrase with a random one from the same intent.
    \item \gc{\textbf{Itemname}}\footnote{With \textit{itemname} we indicate the product mention, \eg \textit{the latest iphone} in the sentence \textit{I want to buy the latest iphone}.} \gc{\textbf{Specificity Manipulation (ISM)}} (only for SUG): we modify an increasing number of texts in $R$ by making their itemname broader (removing product attributes identified by a BERT-based NER model \cite{filice-etal-2021-voiser}) or more specific (adding popular attributes associated to the product).
\end{itemize}

We use 3,833, 2,085 and 5,000 synthetic rankings $\mathcal{R}^*$ for SUG, PQG and QAC, respectively. Each contains one reference bag and a sequence of 5 manipulated bags, ranked by their level of noise. 
For efficiency, we limit each bag size to 100 by randomly down-sampling larger bags. Reference bags contain 32.07, 14.78 and 9.59 texts, with a 34.40, 21.99 and 1.19 standard deviation, for SUG, PQG and QAC, respectively. 

\begin{figure}
  \includegraphics[scale=0.49]{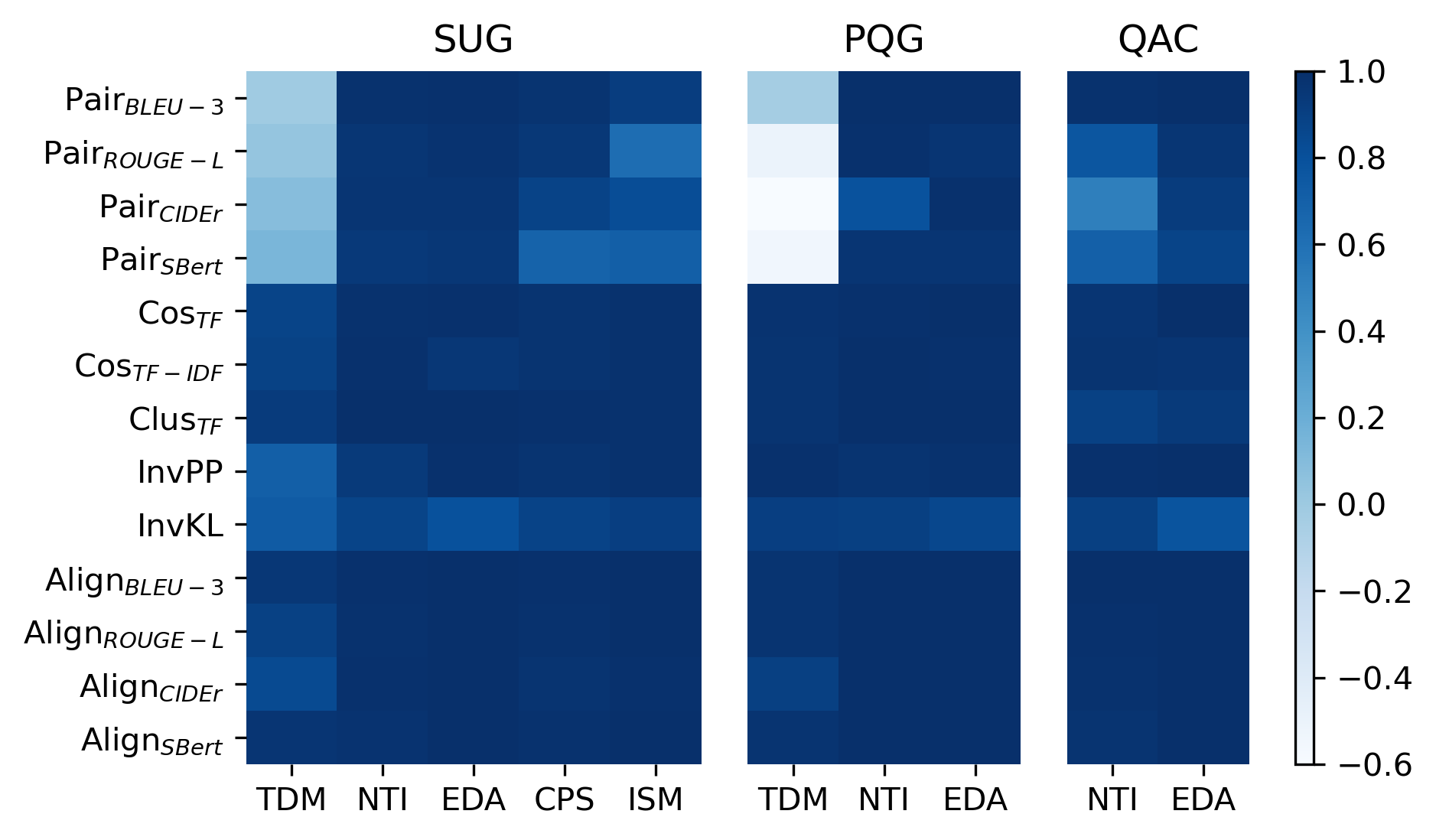}
  \caption{Spearman correlation between the real rankings and the predicted rankings for different metrics.}
  \label{fig:man_results}
\end{figure}

Figure \ref{fig:man_results} reports the Spearman correlation between the real rankings $\mathcal{R}^*$ and the ones induced by different metrics $\mathcal{R}_m$.
We can observe that, pairwise metrics perform poor on TDM: the correlations are very low in SUG and PQC. Compared to other metrics, they exhibit low correlations also on NTI, especially in the QAC task. 
When the alignment is applied to pairwise metrics (both lexical and learned), we observe significant increases in correlation in all cases, suggesting the effectiveness of the proposed alignment. We argue that aggregating every possible pairwise scores favor the head of the distribution (\ie frequent expressions/terms), while finding the optimal alignment better considers also the tail.

On almost all manipulations, Pair$_{\text{SBert}}$ performs worse than Pair$_{\text{BLEU-3}}$, and similarly Align$_{\text{BLEU-3}}$ achieves better correlations than Align$_{\text{SBert}}$. We suspect pre-trained models are not calibrated enough to evaluate texts from $R$ that share extremely similar lexical patterns (\ie utterances with same intent-product pair in SUG or auto-completed queries from QAC). 

Document metrics (\ie Cos$_{\text{TF}}$ or Cos$_{\text{TF-IDF}}$) show strong and consistent performances in all three tasks. This is because representing an entire bag with a single representation preserves the word distribution of the bag for both tail and head expressions.
Lastly, InvPP, InvKL and Clus$_{\text{TF}}$ are also competitive metrics.
 
 \begin{figure*}[!ht]
     \centering
     \begin{subfigure}[b]{0.327\textwidth}
         \centering
         \includegraphics[width=\textwidth]{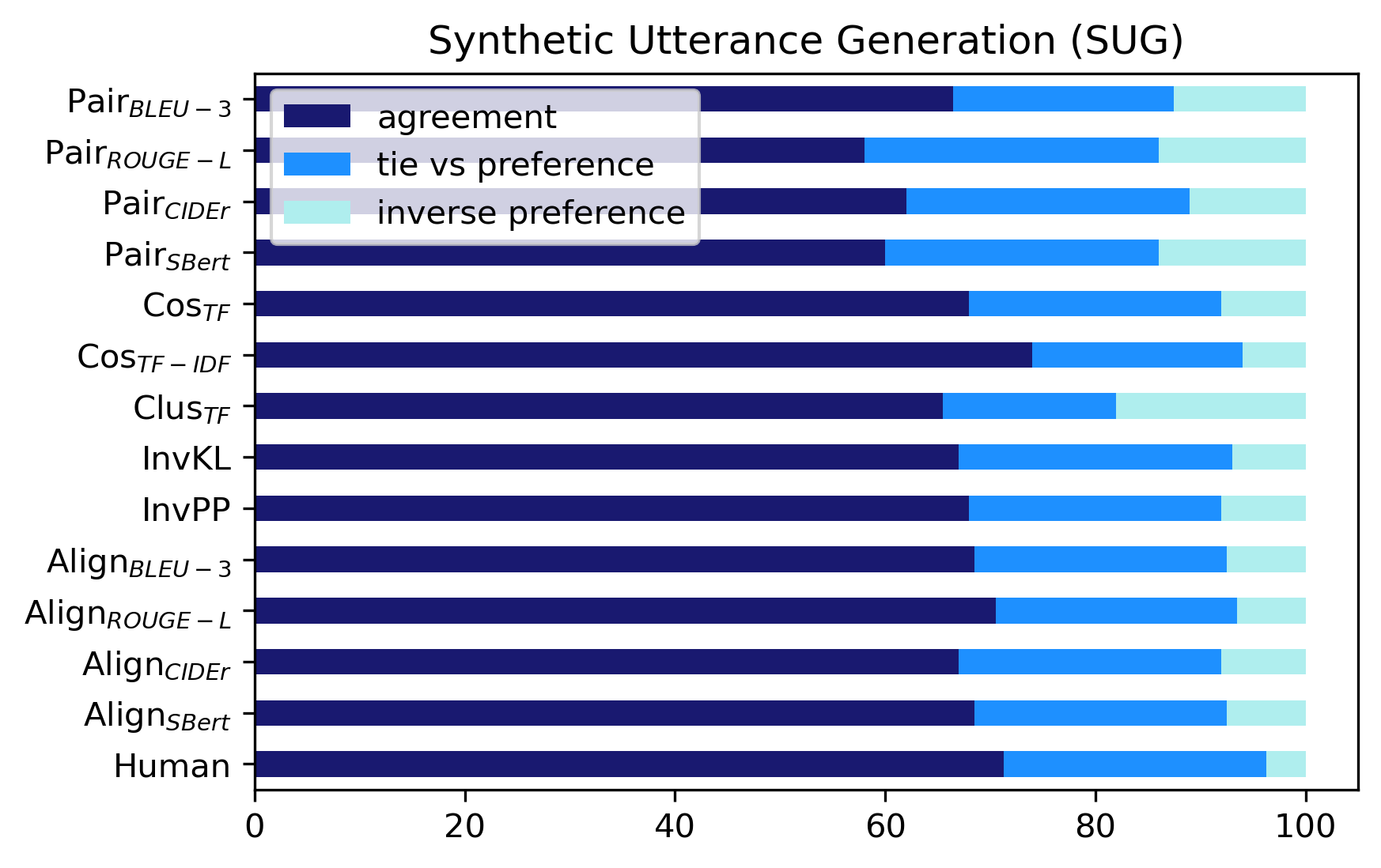}
     \end{subfigure}
     \begin{subfigure}[b]{0.327\textwidth}
         \centering
         \includegraphics[width=\textwidth]{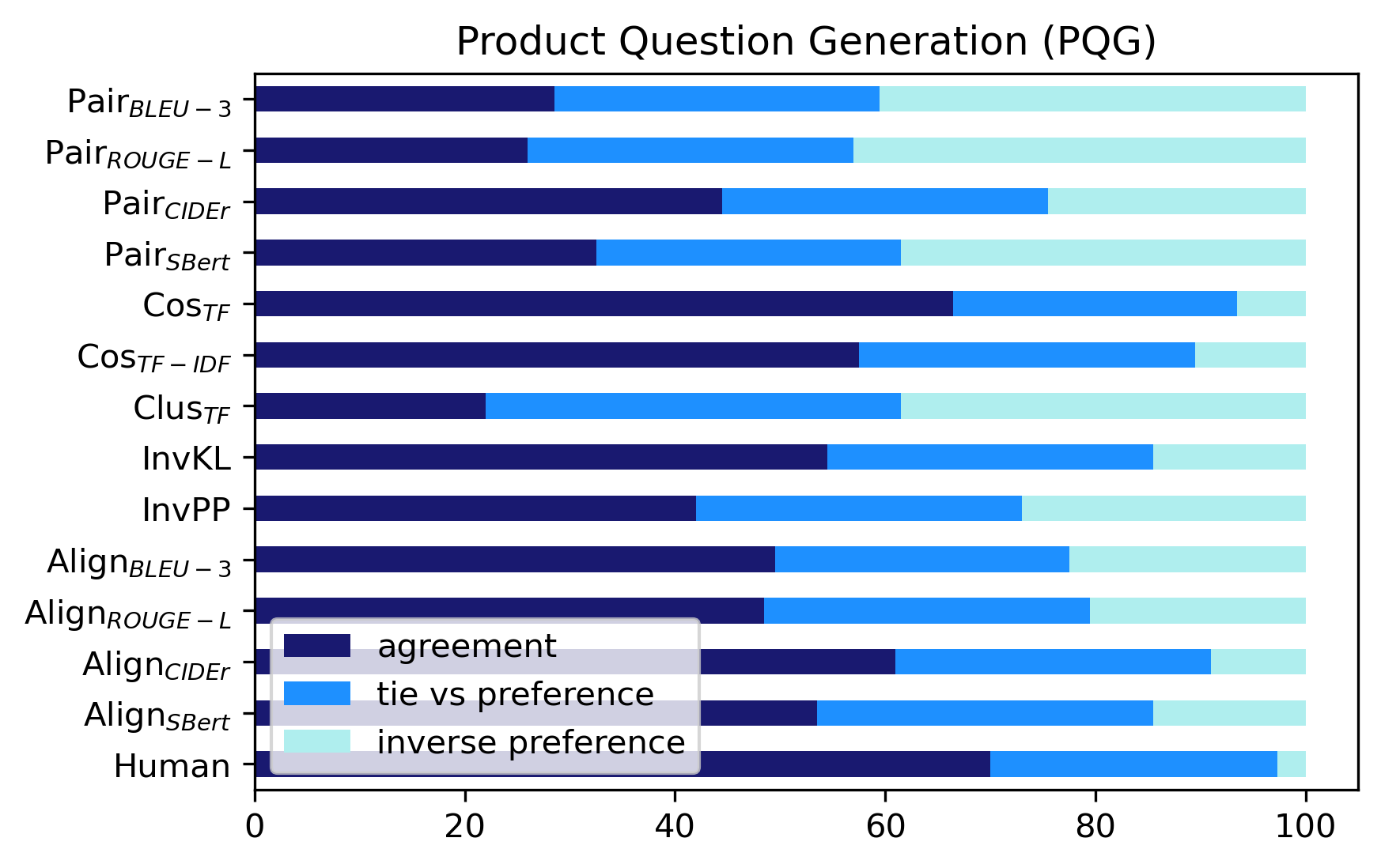}
     \end{subfigure}
     \begin{subfigure}[b]{0.327\textwidth}
         \centering
         \includegraphics[width=\textwidth]{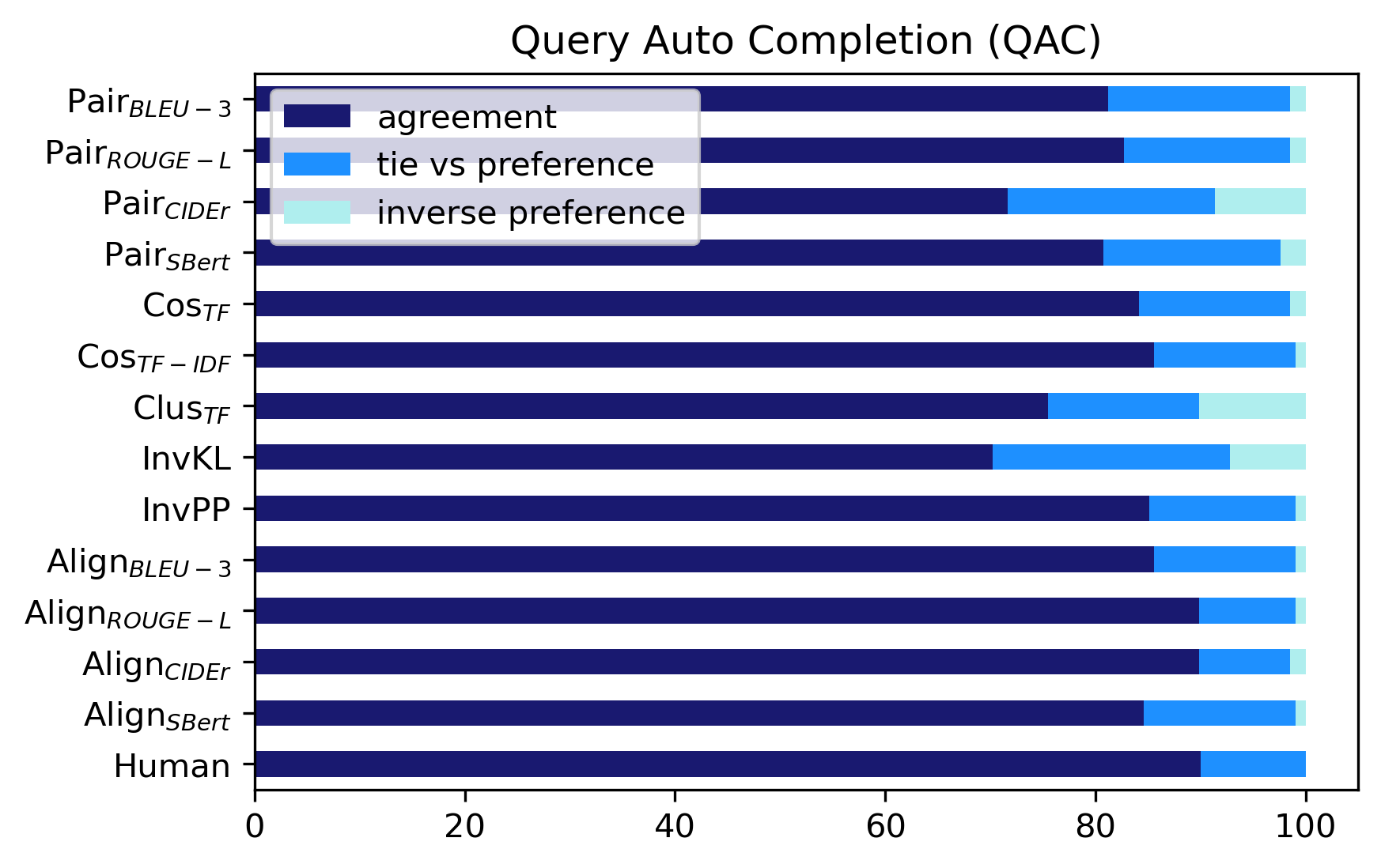}
     \end{subfigure}
        \caption{Results of the human annotations. \texttt{agreement}: annotators agree with our metric. \texttt{tie vs preference}: annotators select the equally good choice and the difference between the metric scores is above the tie threshold. \texttt{inverse preference}: annotators prefer the bag scored the least by our metric. }
\label{fig:ann_results}
\end{figure*}

\begin{figure}
  \centering
  \includegraphics[width=\linewidth]{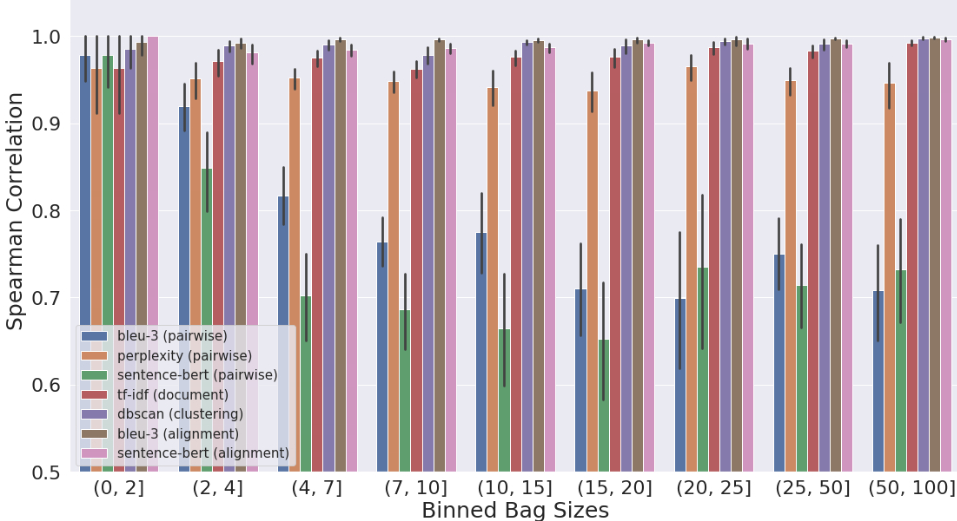}
  \caption{Analysis on comparing Spearman Correlation by different bag sizes on SUG dataset.}
  \label{fig:bag_size_plot}
\end{figure}

\noindent
\textbf{Rank Correlation vs. Bag Sizes}. We further study how different metrics perform across different bag sizes. We select the top performing 7 metrics and we focus on SUG, as in PQG $R$ contains on average less than 5 questions and is less comprehensive compared to SUG, while in QAC the reference bag always contains 10 auto-completions.
As shown in Figure \ref{fig:bag_size_plot},  pairwise metrics suffer from performance loss as bag size increases. For instance, pairwise-BLEU-3 starts with almost perfect correlation (1.0), and degrades to ~0.75 for bag sizes > 50. The trend is similar for Sentence-BERT, but the drops are much more significant. Conversely, when alignment is applied to pairwise metrics, performances are consistently strong across all bag sizes. It seems that alignment significantly reduces noise by finding the maximal alignment among two bags. 
For document and clustering-based metrics, there is a slight increase in performance as the bag size increases. Theoretically, document metrics should perform stronger with larger bags. However it was surprising to see that these metrics perform almost equally well on smaller bag sizes (e.g. size <= 2). For TF-IDF approaches, this makes sense because individual sentence vectors are computed first and summed to represent the bag. Hence, each sentence encoding still carries its meaning. 

\subsection{Human Evaluation}
\label{sec:human_evaluation}

 \begin{figure*}[!ht]
 \centering
  \includegraphics[angle=270, width=0.9\linewidth]{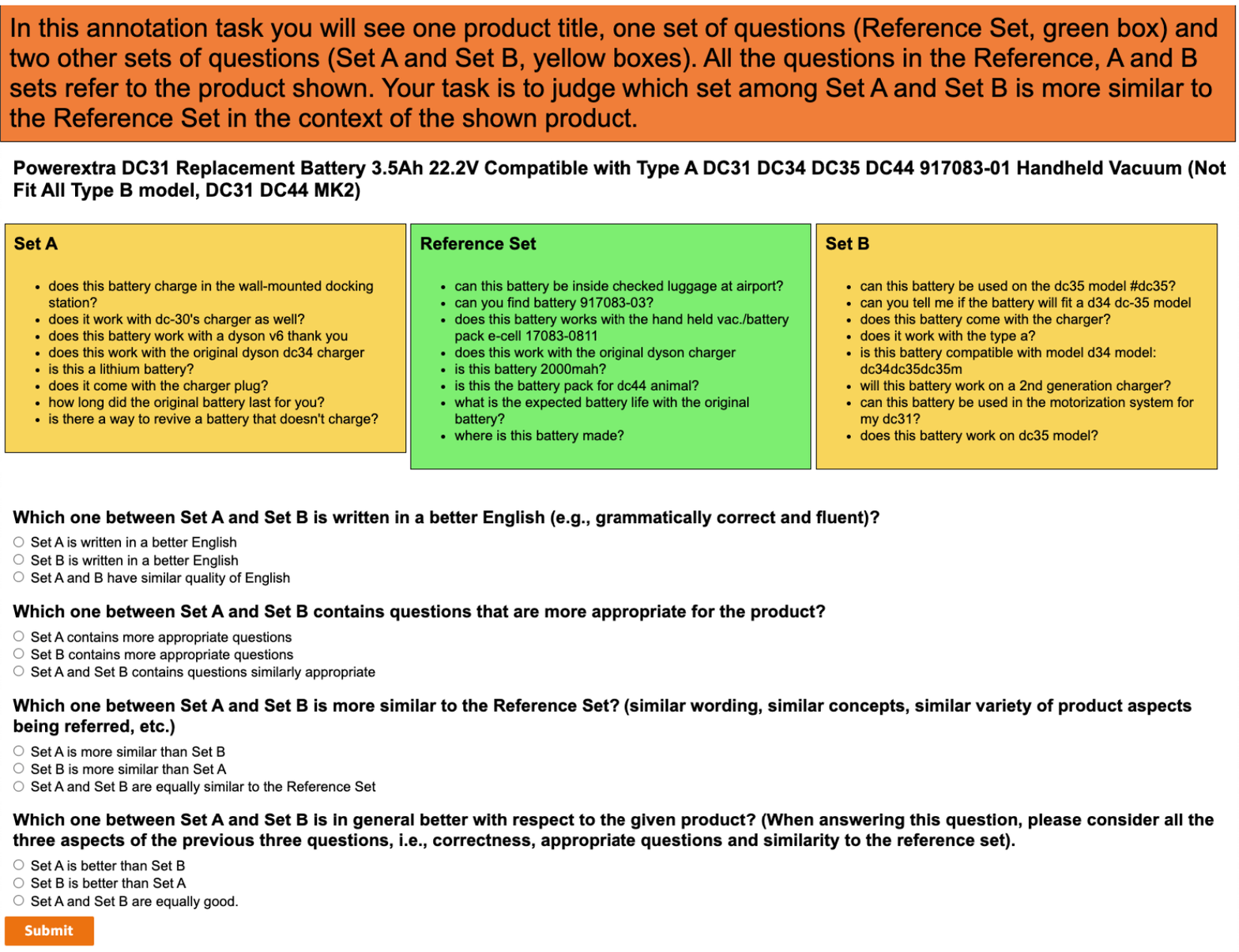}
  \caption{Product Question Generation annotation task example.}
  \label{fig:pqg_annotation}
\end{figure*}

The generated bags we use in the automatic procedure are synthetically obtained by manipulating the reference bag, and might not fully resemble the real quality issues introduced by NLG models. Thus, we also run a human annotation task on bags generated by NLG models, and ask human experts to rate them.\\

\noindent
\textbf{Annotation Task}. We opt for a comparative annotation task, where annotators provide their preference between two generated bags; the comparative approach helps reducing subjectivity and typically leads to better annotators' agreement \cite{callison-burch-etal-2007-meta}. 
Figure \ref{fig:pqg_annotation} illustrates an example of our annotation task for PQG. Annotators are given the following information:
\begin{itemize}
    \item \textit{Context}: In SUG, the context is made of the product title and the intent. In PQG, the context is the product title. In QAC the context is the web query.
    \item \textit{Reference Bag}: a bag of texts containing the reference data related to the shown context.
    \item \textit{Generated Bags}: two bags of texts generated with two different models.
\end{itemize}

In each annotation task we collect preferences on: Q$_1$ fluency and grammatical correctness; Q$_2$ relevancy to the context (the product in SUG and PQG and the query in QAC); Q$_3$ similarity to the reference bag; Q$_4$ overall preference. Our analysis considers only Q$_4$, but the other questions are useful to let the annotators focus on different quality aspects before expressing their overall preference. Human experts (\ie full-time scientists) annotated 200 bag pairs for each task. A subset of these pairs were annotated by multiple annotators and we measured a satisfactory agreement on Q$_4$: Fleiss Kappa 0.437 in SUG, 0.537 in PQG, and 0.824 on QAC. Most of the disagreement (see last bar in Figure \ref{fig:ann_results}) occurs when one annotator expresses a tie, while the other a preference. This is a non-severe error which can happen when an annotator notices a difference that the other does not observe or judges as marginal.\\

\noindent
\textbf{Traffic Generation Models}.
For PQG, we consider the following models: \Ni BART-base \cite{DBLP:journals/corr/abs-1910-13461} with beam search (\texttt{beam-size}=10); \Nii BART-large with nucleus sampling \cite{DBLP:journals/corr/abs-1904-09751} (\texttt{top-p}=0.9). For SUG, we use: \Ni a template-based solution where predefined intent-related carrier phrases are combined with itemnames extracted from product titles; \Nii a BART-base model with nucleus sampling (\texttt{top-p}=0.9). For QAC we consider \Ni BART-base with beam search (\texttt{beam-size}=10) and \Nii T5-base \cite{DBLP:journals/corr/abs-1910-10683} with nucleus sampling (\texttt{top-p}=0.9).
We trained all the models for 15 epochs and applying the Early Stopping with patience 3. We limited the maximum length to 256. For BART-base and T5-base we adopted a batch size 32, while for BART-large the batch size was set to 8, due to memory limitations. All the models have been acquired with 4 Nvidia V100 GPUs.
In all tasks, we also consider real texts as one of the bags under comparison: this bag and the reference bag are two different samples from the same distribution, \ie utterances about the same intent-product in SUG, questions about the same product-aspect in PQG and top query auto-completions in QAC.\\

\noindent
\textbf{Metric-to-Human Correlations}.
Figure \ref{fig:ann_results} reports the metric-to-human agreement, measured in accuracy. 
For each metric we estimate a similarity threshold to express ties: if the difference between the metric scores assigned to two bags is below the threshold, we consider the bags equally good. The threshold is set so that the percentage of ties equal the number of ties expressed by humans (16.5\% for SUG, 20.0\% for PQG, and 15.1\% for QAC). 
Human evaluation confirms that the na\"ive usage of sentence similarity metrics (\ie the pairwise metrics) is not effective to measure the quality of generated traffic, while the application of the alignment strategy yields substantial improvements. 


For all tasks, document metrics (in particular, Cos$_{\text{TF-IDF}}$ for SUG and QAC and Cos$_{\text{TF}}$ for PQG) perform very consistently and are comparable to inter-annotator agreement, \ie they select the best bag (or correctly identify a tie). In SUG, inverse document frequency (IDF) helps to focus on itemnames rather than carrier phrases (which have a pretty limited vocabulary). Similarly, in QAC IDF helps to focus on terms that are not part of the original input query. Also annotators privilege this novel terms when evaluating bag similarity, giving Cos$_{\text{TF-IDF}}$ some advantage.
On the other side, in PQG there are many rare words (\eg numerical tokens related to product models, dimensions, etc.) and when IDF assigns high weights on them, performance degrades.
Higher results on QAC can be justified by its simpler evaluation setting: the bags contain distinct queries and this emphasizes their differences, making their comparison easier for both humans and metrics.

Overall, we argue that evaluating STG models cannot be done with standard metrics; instead, we need to consider the generated traffic as a whole. We claim that computing bag-level representations using document metrics (\ie Cos$_{TF-IDF}$ or Cos$_{TF}$) produce the most consistent solution, especially on tasks that require to generate texts with different prevalence, like SUG and PQG where we observe 8\% and 22\% agreement improvement w.r.t. the best pairwise metric. 
\vspace{-0.25em}
\section{Conclusions}
\label{sec:conclusions}

This paper introduced the Synthetic Traffic Generation task, which requires a single-prompt-multi-output NLG solution, and its importance in real-world applications (\eg for conversational agents). We tested the applicability of standard NLG evaluation metrics, like BLEU, that individually judge the quality of the generated utterances. Through extensive evaluations on publicly available and industry datasets, we observed that \gc{standard NLG metrics do not capture all the nuances of a distribution of texts. We proposed metrics that consider the generated traffic as a whole.
In our experiments, document-based metrics, where we represent a text distribution as a single vector (\eg a TF or TF-IDF representation), which can be compared to other distributions through cosine similarity, provides the most consistent solution.} 
On tasks requiring to generated a full text distribution we observed up to 20\% metric-to-human correlation improvement w.r.t. standard NLG metrics. 
\simo{While further work is needed to define better strategies for evaluating whether synthetic traffic is representative, we believe that our work provides a good starting point.} These findings can help reducing the need for human annotations by supporting the development of better Synthetic Traffic Generation models. In fact, these models can be used to produce realistic data for optimizing or testing NLP pipelines in conversational agents.





\clearpage

\bibliography{main}

\end{document}